\documentclass{article}

\usepackage[preprint]{neurips_2025}

\usepackage[utf8]{inputenc} 
\usepackage[T1]{fontenc}    
\usepackage{hyperref}       
\usepackage{url}            
\usepackage{booktabs}       
\usepackage{amsfonts}       
\usepackage{nicefrac}       
\usepackage{microtype}      
\usepackage{xcolor}         
\usepackage{graphicx}
\usepackage{amsmath}
\usepackage{multirow}

\title{OpenGT: A Comprehensive Benchmark For Graph Transformers}

\author{%
  \textbf{Jiachen Tang}$^1$,
  \textbf{Zhonghao Wang}$^{2}$,
  \textbf{Sirui Chen}$^1$, 
  \textbf{Sheng Zhou}$^{2}$, \\
  \textbf{Jiawei Chen}$^1$, 
  \textbf{Jiajun Bu}$^1$\\
  $^1$Zhejiang Key Laboratory of Accessible Perception and Intelligent Systems, \\
  College of Computer Science,  Zhejiang University\quad \\
  $^2$ School of Software Technology,  Zhejiang University\quad \\ \texttt{
  \{tangjc, wangzhonghao, chenthree, zhousheng\_zju, }\\
  \texttt{sleepyhunt, bjj\}@zju.edu.cn}
}

\begin{document}

\maketitle

\begin{abstract}
  Graph Transformers (GTs) have recently demonstrated remarkable performance across diverse domains.  
  By leveraging attention mechanisms, GTs are capable of modeling long-range dependencies and complex structural relationships beyond local neighborhoods. However, their applicable scenarios are still underexplored, this highlights the need to identify when and why they excel. 
  Furthermore, unlike GNNs, which predominantly rely on message-passing mechanisms, GTs exhibit a diverse design space in areas such as positional encoding, attention mechanisms, and graph-specific adaptations. Yet, it remains unclear which of these design choices are truly effective and under what conditions. 
  As a result, the community currently lacks a comprehensive benchmark and library to promote a deeper understanding and further development of GTs. 
  To address this gap, this paper introduces OpenGT, a comprehensive benchmark for Graph Transformers. OpenGT enables fair comparisons and multidimensional analysis by establishing standardized experimental settings and incorporating a broad selection of state-of-the-art GNN and GTs. Our benchmark 
  evaluates GTs from multiple perspectives, encompassing diverse tasks (both node-level and graph-level) and datasets with varying properties, such as scale, homophily, and sparsity.
  Through extensive experiments, 
  our benchmark has uncovered several critical insights, including the difficulty of transferring models across task levels, the limitations of local attention, the efficiency trade-offs in several models, the application scenarios of specific positional encodings, and the preprocessing overhead of some positional encodings.
  We aspire for this work to establish a foundation for future graph transformer research emphasizing fairness, reproducibility, and generalizability. 
  We have developed an easy-to-use library OpenGT for training and evaluating existing GTs. 
  The benchmark code is available at \url{https://github.com/eaglelab-zju/OpenGT}.

\end{abstract}

\section{Introduction} 

Recently, Graph Transformers (GTs)\citep{graphtransformer, graphormer, graphgps}, inspired by the self-attention mechanism of the Transformer architecture \citep{vaswani2017attention}, have emerged as promising alternatives to Graph Neural Networks (GNNs)\citep{gcn, gat, appnp} in various graph learning tasks. By enabling flexible, global interactions among nodes, GTs are better positioned to handle a wider range of structural and semantic patterns, and have demonstrated strong performance in both node-level and graph-level tasks. 

Despite the growing body of work on GTs, the field lacks a unified and extensible framework for implementing, comparing, and analyzing these models in a standardized setting. 
Existing studies often vary significantly in terms of experimental settings, datasets, positional encodings, and architectural components, making it difficult to draw fair or generalizable conclusions \citep{shchur2018pitfalls, hu2020open, dwivedi2022long}. 
Furthermore, unlike GNNs, which predominantly rely on message-passing mechanisms, GTs exhibit a highly diverse design space. The diversity of graph structure encoding strategies~\citep{survey}, positional encodings~\citep{benchmarkpe}, and hybrid architectural designs has resulted in a fragmented landscape, where performance comparisons are often inconsistent or incomplete. 

\begin{figure}
    \centering
    \includegraphics[width=\linewidth]{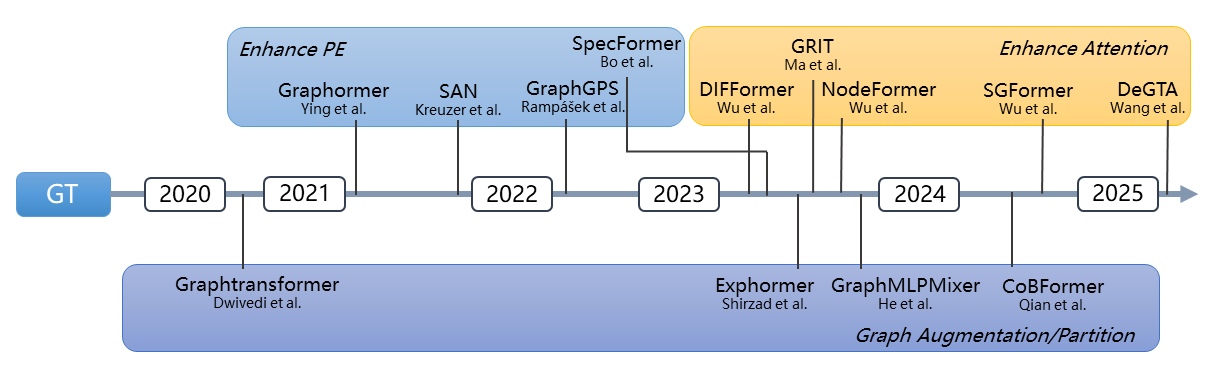}
    \caption{Timeline of Graph Transformer development. Graph Transformers are categorized according to their key improvement.}
    \label{fig:timeline}
\end{figure}

To address this gap, our work provides a unified and reproducible benchmark for evaluating a diverse set of graph learning models using the modular \texttt{torch\_geometric.graphgym} framework \citep{graphgym}. We re-implement a wide range of GNNs and GTs under a consistent design and training pipeline, spanning multiple datasets and task types. Figure \ref{fig:timeline} shows a timeline of implemented GTs. We categorize these models into three categories according to their innovation in positional encodings, attention mechanisms, and graph preprocessing methods. Our evaluation includes systematic studies on model performance, architectural design choices, positional encodings, and computational efficiency. 

Our contributions can be summarized as follows:
\begin{itemize}
    \item \textbf{Unified Benchmark Library.} We release an open-source benchmark library, implementing 16 representative graph transformer and GNN models. The library incorporates a diverse suite of datasets covering both node-level and graph-level tasks, with variations in size, sparsity, and homophily.

    \item \textbf{Systematic Evaluation.} We conduct extensive experiments to evaluate existing graph transformer models across multiple axes, including task type (node vs. graph level), graph structure (homophilic vs. heterophilic), dataset scale, attention mechanisms, and integration strategies of graph-specific information.

    \item \textbf{Empirical Observations.} Through controlled and reproducible experiments, we derive several key findings regarding the effectiveness, limitations, and computational behavior of current graph transformer designs. These insights include the role of positional encodings, architectural components, and the trade-offs between accuracy and efficiency.

    \item \textbf{Future Research Directions.} Based on our findings, we outline actionable suggestions for future graph transformer research, such as improving the scalability of positional encodings, developing effective strategies for selecting positional encodings, and exploring efficient partition-based attention mechanisms.
\end{itemize}


\section{Preliminaries}
\textbf{Problem Definition}

Graph structure data is denoted as $\mathcal G=(\mathcal V,\mathcal E)$, where $\mathcal V$ denotes a set of $N$ nodes and $\mathcal E \subseteq \mathcal V \times \mathcal V$ denotes a set of $M$ edges. Each node $v_i \in \mathcal V$ may have associated features $\mathbf x_i \in \mathbb R^d$, forming a feature matrix $\mathbf X \in \mathbb R^{N \times d}$. The goal of graph representation learning is to map nodes, edges, or the entire graph into a low-dimensional embedding to capture the structure and semantic properties of downstream tasks such as node classification, link prediction, and graph classification. 
Specifically, given graph $\mathcal G$ and some known labels $\mathcal Y_{\mathrm{train}}$, the task is to learn a function $f : \mathcal G \rightarrow \hat{\mathcal Y}$. Where $\hat{\mathcal Y}$ depends on the problem setting. In node-level tasks, $\mathcal Y_{\mathrm{train}}$ and $\hat{\mathcal Y}$ correspond to the label of each node; in edge-level tasks, $\mathcal Y_{\mathrm{train}}$ and $\hat{\mathcal Y}$ predict the attributes or existence of the edge; in graph-level tasks, the input contains multiple graphs, and $\mathcal Y_{\mathrm{train}}$ and $\hat{\mathcal Y}$ represent the label of each graph.
Traditional graph neural networks rely on message passing to aggregate local neighborhood information, and often encounter problems such as over-squeezing, over-smoothing, and inability to work with heterophilous graphs. This inspired the exploration of graph transformers, which use global attention mechanisms to overcome these limitations.

\textbf{Graph Transformers}

The Transformer architecture was originally designed for sequential data in the field of natural language processing \cite{vaswani2017attention}. Graph Transformer applies it to graph-structured data by redefining the self-attention mechanism \cite{nodeformer, grit}, introducing positional encoding \cite{graphormer}, and combining it with graph neural networks \cite{difformer,sgformer} to exploit structural and positional information.

The core mechanism of the Transformer is its global self-attention module, which calculates the interactions between all node pairs as follows:
\begin{equation}
  \text{Attn}(\mathbf Q, \mathbf K, \mathbf V)=\text{softmax}(\frac{\mathbf Q\mathbf K^T}{\sqrt{d_K}})\mathbf V 
    \label{attn}
\end{equation}
where the query matrix $\mathbf Q=\mathbf X \mathbf {W}_Q$, the key matrix $\mathbf K=\mathbf X \mathbf {W}_K$, and the value matrix $\mathbf V=\mathbf X \mathbf {W}_V$ are obtained by projecting the original feature matrix $\mathbf X$ through three weight matrices $\mathbf W_Q \in \mathbb R^{d \times d_K}$, $\mathbf W_K \in \mathbb R^{d \times d_K}$, $\mathbf W_V \in \mathbb R^{d \times d_V}$.

Another feature of Transformer is the position encoding designed specifically to identify the absolute and relative positions of entities in a sequence. However, there is no sequence order in the graph, so it is necessary to design a new position encoding $pe$ based on the topological properties of the graph to supplement the missing graph information in the pure Transformer. Different graph transformers use different positional encodings. Some encode node-wise information by adding or concatenating it to node features \cite{graphtransformer,specformer}, i.e., $\hat{\mathbf x}_i=\mathbf x_i+\mathrm{pe}_i$ or $\hat{\mathbf x}_i=[\mathbf x_i,\mathrm{pe}_i]$, and some encode pair-wise information by adding it to the attention weight \cite{san}, i.e., $\text{Attn}(\mathbf q_i,\mathbf k_j, \mathbf v_j)=\sigma(\frac{\mathbf q_i \mathbf k_j^T}{\sqrt{d_K}},pe_{i,j})\mathbf V$.

In addition to using positional encoding to incorporate graph information, some work combines traditional graph neural networks with Transformer models. Additional graph signals are added to the Transformer by alternating between GNN layers and Transformer layers \cite{nodeformer}, or by post-processing and merging the outputs of the two architectures \cite{graphgps,difformer,sgformer}.

The original Transformer shown in Eq\eqref{attn} has a high complexity of $O(N^2d)$ and is not suitable for large graph tasks. Therefore, some methods are usually used to reduce the complexity of Graph Transformers, such as large graph segmentation \cite{cobformer,graphmlpmixer}, low-order neighbor sampling \cite{graphtransformer,graphgps,degta}, etc. In recent years, some work has also attempted to modify the attention mechanism and proposed Transformers with linear complexity \cite{nodeformer,difformer,sgformer}.

\section{Benchmark Design} 

\subsection{Datasets and Implementations}

\textbf{Datasets:} 
We conduct experiments on comprehensive datasets with varying graph sizes, structural characteristics(e.g., homophily, sparsity), and task types for a thorough evaluation of model capabilities across domains, encompassing both node-level and graph-level tasks.
The statistics of the evalauted datasets are illustrated in Table~\ref{tab:dataset-stats}. 

For node-level tasks, we include datasets from various domains. 
The citation networks include \textit{Cora}, \textit{Citeseer}, and \textit{Pubmed} \citep{sen2008collective}, where nodes represent documents and edges represent citation links.
Each document is assigned as single topic as the node classification label.
The webpage networks include \textit{Squirrel} and \textit{Chameleon} \citep{rozemberczki2019gemsec}, which consist of web pages extracted from Wikipedia as nodes in the graph.
These nodes are densely connected via mutual links and labeled by topical domain.
We also include the \textit{Actor} dataset \citep{shchur2018pitfalls}, where nodes represent actors and edges denote co-occurrence on Wikipedia pages, with roles as classification labels.
Finally, we include three datasets derived from university web pages: \textit{Texas}, \textit{Cornell}, and \textit{Wisconsin} \citep{pei2020geom}, where nodes correspond to individual web pages, and edges represent hyperlinks between them. Each node is labeled according to the category of the web page it represents.

For graph-level tasks, we evaluate models on molecular and protein-related datasets. For molecular graphs, we use \textit{ZINC} \citep{irwin2012zinc} for graph regression and two datasets from the Open Graph Benchmark: \textit{OGBG-MolHIV} and \textit{OGBG-MolPCBA} \citep{hu2020open}, which involve binary classification of molecular properties related to drug activity and assay readouts. 
Additionally, we use two peptide datasets from the Long Range Graph Benchmark, \textit{Peptides-func} and \textit{Peptides-struct} \citep{dwivedi2022long}, which consist of peptide molecular structures labeled for functional and structural attributes.

\begin{table}
\centering
\caption{Statistics of the datasets. }
\label{tab:dataset-stats}
\resizebox{\textwidth}{!}{%
\begin{tabular}{lcccccccc}
\toprule
\textbf{Dataset} & \textbf{Level} & \textbf{\#Graphs} & \textbf{Avg. Nodes} & \textbf{Avg. Edges} & \textbf{\#Classes} & \textbf{\#Features} & \textbf{Homophily} & \textbf{Metric} \\
\midrule
Cora             & Node & 1       & 2,708  & 5,429  & 7   & 1,433 & 0.81 & Accuracy \\
Citeseer         & Node & 1       & 3,327  & 4,732  & 6   & 3,703 & 0.74 & Accuracy \\
Pubmed           & Node & 1       & 19,717 & 44,338 & 3   & 500   & 0.80 & Accuracy \\
Squirrel         & Node & 1       & 5,201  & 217,073& 10   & 2,089 & 0.22 & Accuracy \\
Chameleon        & Node & 1       & 2,277  & 31,421 & 10   & 2,325 & 0.23 & Accuracy \\
Actor            & Node & 1       & 7,600  & 30,019 & 5   & 931   & 0.22 & Accuracy \\
Texas            & Node & 1       & 183    & 325    & 5   & 1,703 & 0.11 & Accuracy \\
Cornell          & Node & 1       & 183    & 298    & 5   & 1,703 & 0.30 & Accuracy \\
Wisconsin        & Node & 1       & 251    & 515    & 5   & 1,703 & 0.21 & Accuracy \\
\midrule
ZINC             & Graph     & 12,000  & 23.2   & 49.8   & --  & 28    & --   & MAE \\
OGBG-MolHIV      & Graph & 41,127  & 25.5   & 27.5   & 2   & 9     & --   & ROC-AUC \\
OGBG-MolPCBA     & Graph & 437,929 & 26.0   & 28.1   & 128 & 9     & --   & AP \\
Peptides-Func    & Graph & 15,535  & 151   & 307   & 10  & 9     & --   & AP \\
Peptides-Struct  & Graph & 15,535  & 151   & 307   & 11  & 9     & --   & MAE \\
\bottomrule
\end{tabular}
}
\end{table}

\textbf{Implementation Details}

We implemented all selected models using the \texttt{torch\_geometric.graphgym} \citep{graphgym} framework, which provides a modular and extensible platform for designing and evaluating graph neural networks. Each model is decomposed into standardized modules, including \textit{input encoders, GNN or Transformer layers, graph pooling or readout layers, and task-specific heads}. This modular structure allows different models to share common components while enabling straightforward substitution or extension of architectural elements such as positional encodings, normalization layers, or attention mechanisms. By adhering to this decoupled design, we are able to implement a wide range of models from traditional message-passing GNNs to more recent graph transformers within a unified and reproducible training pipeline. 

Based on the modular structure, we implemented the following graph transformer methods: DIFFormer \citep{difformer}, NodeFormer \citep{nodeformer}, GraphGPS \citep{graphgps}, Graphormer \citep{graphormer}, GRIT \citep{grit}, SGFormer \citep{sgformer}, SAN \citep{san}, SpecFormer \citep{specformer}, Exphormer \citep{exphormer}, Graphtransformer \citep{graphtransformer}, GraphMLPMixer \citep{graphmlpmixer}, CoBFormer \citep{cobformer}, and DeGTA \citep{degta}. 
We also implemented some typical GNNs for comparison and analysis, including GCN \citep{gcn}, GAT \citep{gat}, and APPNP \citep{appnp}. 
Benefit from GraphGym's flexible configuration system for model architecture, data loading, training schedules, and logging, all models are implemented with consistent training and evaluation procedures to ensure fair comparison. 

To ensure consistency and fairness in evaluation, all models are trained using the same data splits and follow identical hyperparameter tuning procedures (See Appendix B). For each graph-level dataset, a fixed batch size is used across all models, although the batch size may vary between different datasets to accommodate their specific characteristics. Each experiment is repeated three times with different random seeds, and we report the average performance along with standard deviations to account for variability arising from random initialization and system-level noise. All timing results are measured on an NVIDIA GeForce RTX 3090 GPU with 24~GB of memory.


\subsection{Research Questions}

\textbf{RQ1: Does Graph Transformers outperform Graph Neural Networks with message passing?}

\textbf{Motivation:} 
Graph Transformers and GNNs are the leading techniques in graph mining, each having demonstrated success. However, their respective strengths and weaknesses in various scenarios lack systematic investigation. Understanding these differences is crucial to delineate their specific use cases and limitations.

\textbf{Experiment Design:}
We conduct a series of experiments to compare the two model classes under diverse graph learning scenarios. The evaluation covers a wide range of dataset characteristics to ensure a comprehensive and balanced evaluation.
Specifically, we include \textit{both graph-level and node-level prediction tasks}, representing different granularities of learning objectives. 
For each task type, we select datasets that vary along two key dimensions: \textit{Homophily vs. Heterophily} and \textit{Graph Scale (Small vs. Large)}.
By incorporating datasets across these axes, the experimental setting is structured to provide a broad and systematic comparison of GTs and GNNs based on message passing schema.

\textbf{RQ2: How does the attention mechanism affect the GTs?}

\textbf{Motivation:} 
The attention mechanism has played crucial role in transformer-based methods.
Among existing GTs, some have utilized local attention mechanism by calculating attention score between directly connected nodes. Others calculate between distant nodes. We name the two way of attention calculation as \textit{local attention} and \textit{global attention}.
Different from GNNs where the information of distant neighborhoods can be propagated by message passing, the choice of attention mechanism in GTs have changed the way of information propagation.

\textbf{Experiment Design:}
To investigate how the scope of attention affects model performance, we categorize Graph Transformer models into three groups based on their use of local and global attention mechanisms. Specifically, \textit{Graphtransformer} and \textit{GraphMLPMixer} employs purely local attention. In contrast, \textit{Graphormer}, \textit{NodeFormer}, and \textit{GRIT} use purely global attention, allowing interactions between all node pairs regardless of connectivity. The remaining models incorporate a hybrid approach, combining both local and global attention mechanisms. We conduct experiments on all datasets and compare the performance of models across these categories, aiming to understand the strengths and weaknesses associated with each attention design choice under different graph properties.

\textbf{RQ3: How do Graph Transformers achieve effectiveness-efficiency tradeoff?}

\textbf{Motivation:} 
Although Graph Transformers have demonstrated promising performance and potential, their attention mechanisms typically require computing pairwise weights between nodes in the graph. As a result, early Graph Transformers were often limited to datasets with relatively small graph sizes. Recent studies have attempted to address this limitation; however, the trade-off between effectiveness and efficiency in current Graph Transformer models remains insufficiently understood.

\textbf{Experiment Design:}
To answer this research question, we conduct a series of controlled experiments focused on measuring and analyzing the computational resource requirements of representative models.
Our evaluation focuses on the time efficiency (\textit{training time before reaching best validation performance}),  of the models. 
Note that we will analyze preprocessing time in the next experiment, so we will temporarily ignore it here.
Experiments are conducted on a representative subset of datasets from the earlier sections, selected to include both small and large graphs, as well as node-level and graph-level tasks. 




\textbf{RQ4: How to design positional encodings for Graph Transformers in different scenarios?}

\textbf{Motivation:}
Positional encoding\citep{positional_encoding_1, positional_encoding_2} has been shown to play a crucial role in the effectiveness of Transformer-based models for different domains. Similarly, in Graph Transformers, different datasets, tasks, and domains often require distinct forms of positional encoding. However, due to the diversity of graph data, there has not yet been a systematic exploration or consensus on how to design appropriate positional encodings for different scenarios. This lack of guidance significantly limits the practical application of Graph Transformers in real-world settings.

\textbf{Experiment Design:}
To evaluate the contribution of different positional encodings to model performance in different graph learning tasks, we select multiple models as testbeds to replace their positional encodings.
Here, we categorize positional encodings into Spatial-based Encodings and Spectral-based Encodings according to the type of graph information they provide, and examine the effects of introducing these encodings. 
\begin{enumerate}
    \item \textbf{Spatial-based Encodings} capture the position or distance of nodes in the graph relative to other nodes, such as Graphormer Bias Encoding (encoding pairwise distances and node degrees on the graph), Graph Isomorphism Tokens (inspired by the Weisfeiler-Lehman test, iteratively updating node representations based on neighborhood multisets and acting as structural fingerprints), and Random Walk based Encodings (capturing the probabilistic proximity between nodes through the transition matrix). 
    \item \textbf{Spectral-based Encodings} are derived from the spectral decomposition of graph operators, the most common of which is to use the Laplacian Eigenvalues and Eigenvectors of the graph. Variants include the equivstable version used by Exphormer model\citep{exphormer,eslappe}.
\end{enumerate}
From each type of positional encoding, we select at least one and integrate it into GraphGPS, DIFFormer, and SGFormer through a consistent mechanism, with the underlying model architecture remaining unchanged during the experiments. Here, we select several datasets that ensure that all positional encodings can run successfully within reasonable time and memory constraints, and the selected datasets differ in structural properties such as size and homophily. The above settings allow us to evaluate under which conditions and dataset characteristics positional encodings are most beneficial, and which specific encodings offer the greatest improvement in model performance.

\section{Experiment Results and Analyses}

\subsection{Performance Comparison} 

\begin{figure}
    \centering
    \includegraphics[width=1 \linewidth]{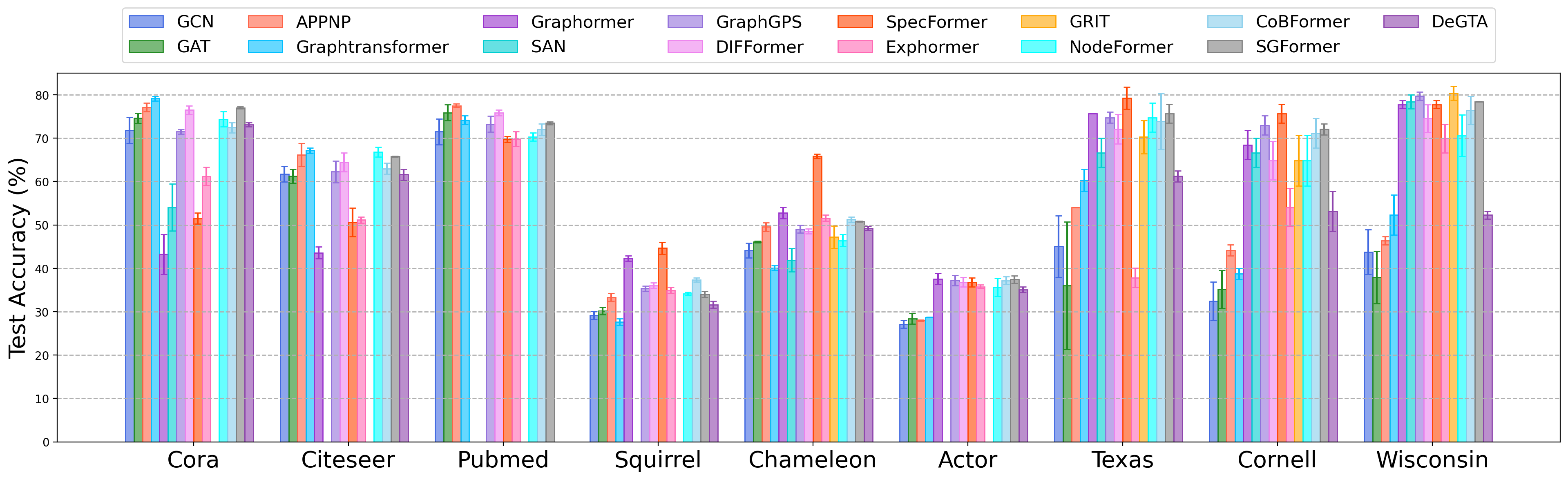}
    \caption{The performance overview on node-level tasks. The empty bar denotes that the model cannot run on the evaluated hardware, rather than achieving zero accuracy.}
    \label{results_acc}
\end{figure}

\begin{figure}
    \centering
    \includegraphics[width=\linewidth]{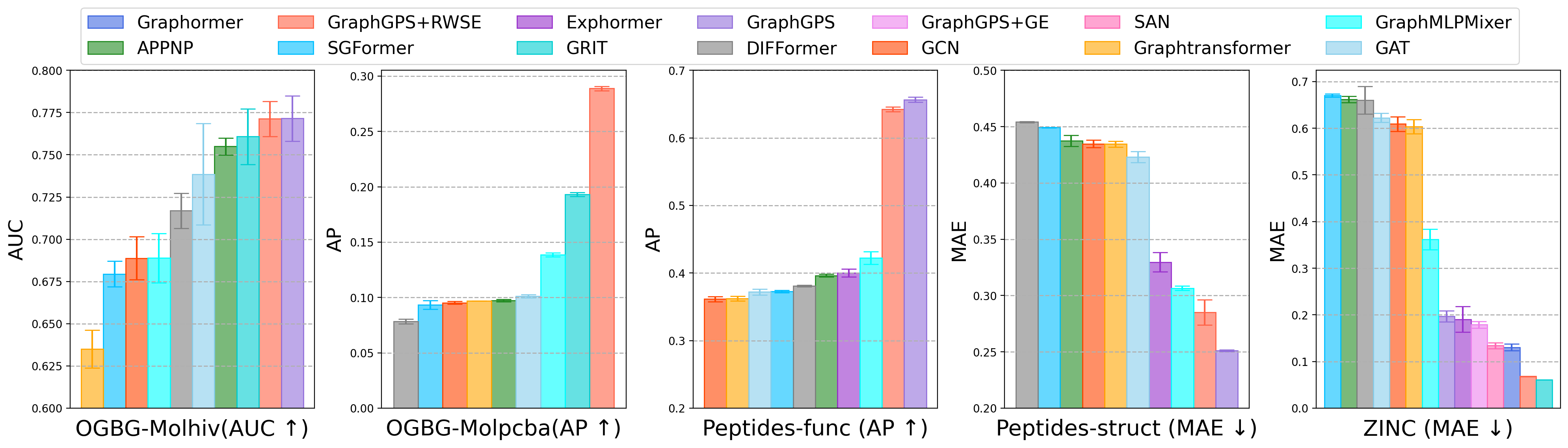}
    \caption{The performance overview on graph-level tasks. The metric used by each dataset is written beside the dataset name, with an arrow showing better performance direction. Note that the missing bar denotes that the model cannot run on the evaluated hardware.}
    \label{results_graph}
\end{figure}

\textbf{Observation 1: Graph transformers achieves more competitive results on heterophilous Graphs.} 
As illustrated in Figure~\ref{results_acc}, in homophily datasets, many GTs and GNNs achieve compariable performance with classic GNNs. 
Meanwhile, we also observe a variance on the GTs where some GTs perform poor or even fail to run on the evaluated hardware. 
This motivates us further explore the design space of the GTs in the future works.
However, on heterophilous graphs, we can observe that 
most GTs significantly outperform compared GNNs.
This can be attributed to the attention mechanisms and flexible receptive fields in GTs, which are particularly effective in settings where neighboring nodes are not necessarily similar. 
Note that we do not discuss the heterophily GNNs which are specialized designed for heterophily graphs. Here we treat the vanilla GNNs as important baselines to verify the GTs on both homophily and heterophily datasets, rather than seeking the best model on heterophily graphs.
Although we do not observe a consistent results on whether GTs can outperform GNNs, the difference between homophily and heterophily datasets suggest that GTs are more flexible when the neighborhood patterns are complex or unknown.

\textbf{Observation 2: Transferability of GTs across task types.} 
In our experiments on graph-level tasks (Figure ~\ref{results_graph}), we observe that Graph Transformer models originally designed for both graph-level and node-level settings—particularly GraphGPS+RWSE and GRIT, which incorporates random walk based positional encodings—tend to outperform traditional GNNs. 
In contrast, GTs primarily designed for node-level tasks do not consistently surpass GNNs when applied to graph-level datasets. 
This suggests that architectural components effective for node-level prediction do not readily generalize to graph-level learning, highlighting the challenge of designing versatile GT architectures that perform well across different task types.
Although this observation is acceptable, the success of GTs in other domains (e.g. vision, language etc.) where different task types can be naturally transferred has raise an expection on the graph domain. This is critical in graph mining, especially on the graph foundation model scenario~\citep{liu2023towards}.

\subsection{Attention Scope Comparison}

\textbf{Observation 3: Limitations of local attention in sparse graphs.} 
Compare among GTs with local and global attention, we can observe that models mainly depending on local attention mechanisms, such as Graphtransformer and DeGTA, tend to exhibit performance degradation on small, heterophilous, and sparse graphs. 
Meanwhile, GTs utilizing global attention (e.g. Graphormer and GRIT) and combing local-global attention (e.g. GraphGPS and SGFormer) maintain competitive performance on these graphs.
This suggests that relying solely on local neighborhood information may be insufficient in such settings, where limited connectivity restricts the model's ability to capture broader contextual information. 
The lack of global or long-range message passing in these architectures appears to constrain their effectiveness on sparse topologies.
Although global attention seems more applicable, we would like to highlight that the global attention also meets the problem of over-globalization problem~\citep{cobformer}. This poses opportunities of balancing the effectiveness and long-range information propagation.




\begin{figure}[htbp]
    \centering
    \includegraphics[scale=0.595]{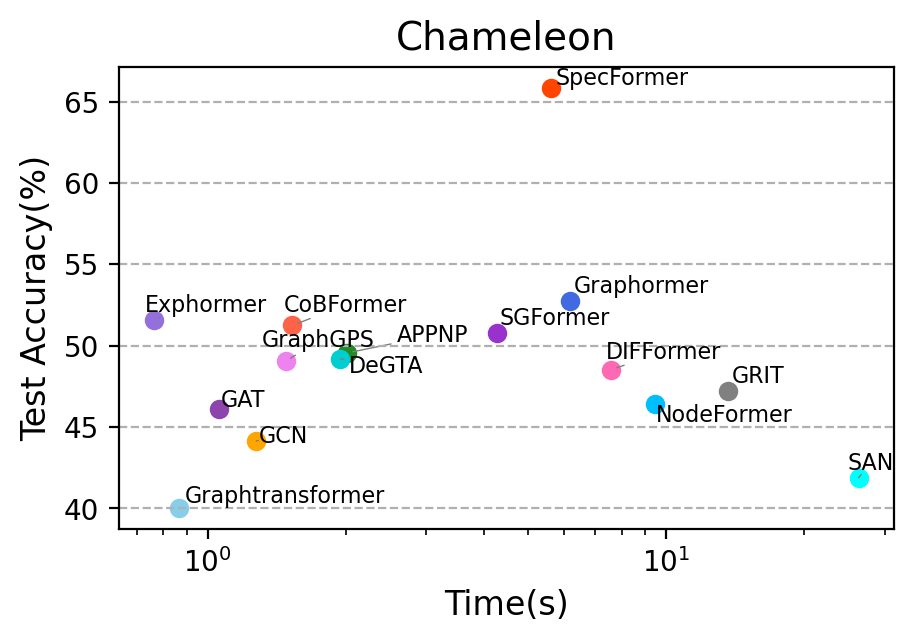}
    \includegraphics[scale=0.595]{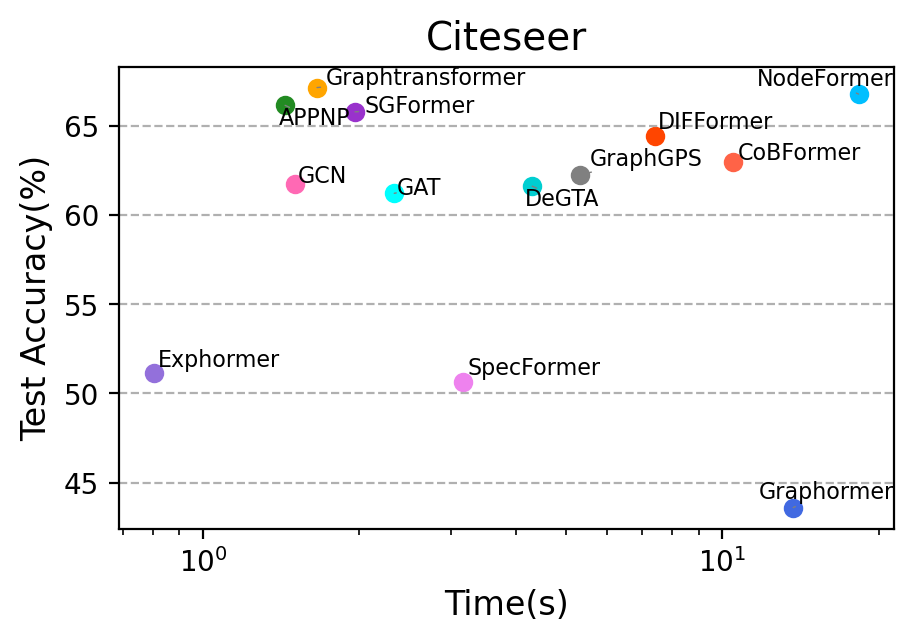}
    \caption{Time consumption and Test accuracy of different methods on Chameleon and Citeseer datasets}
    \label{time}
\end{figure}



\subsection{Time Efficiency}


\textbf{Observation 4: Graph partition is useful for achieving the effectiveness-efficiency tradeoff.}
From Figure~\ref{time}, we can observe that GTs utilizing graph partition exhibit a promising trade-off between accuracy and efficiency. 
In particular, \textit{CoBFormer} for node-level tasks and \textit{GraphMLPMixer} for graph-level tasks achieve competitive performance while incurring lower computational overhead compared to full-graph attention models. 
These models operate by dividing the input graph into partitions or blocks and applying both intra-block and inter-block attention mechanisms. 
This design reduces the complexity of global attention by structuring the computation hierarchically, enabling efficient information exchange across the graph without incurring the full cost of dense all-pairs interactions. 
This suggests an active research direction in achieving the tradeoff from the data perspective.

\begin{figure}
    \centering
    \includegraphics[width=1\linewidth]{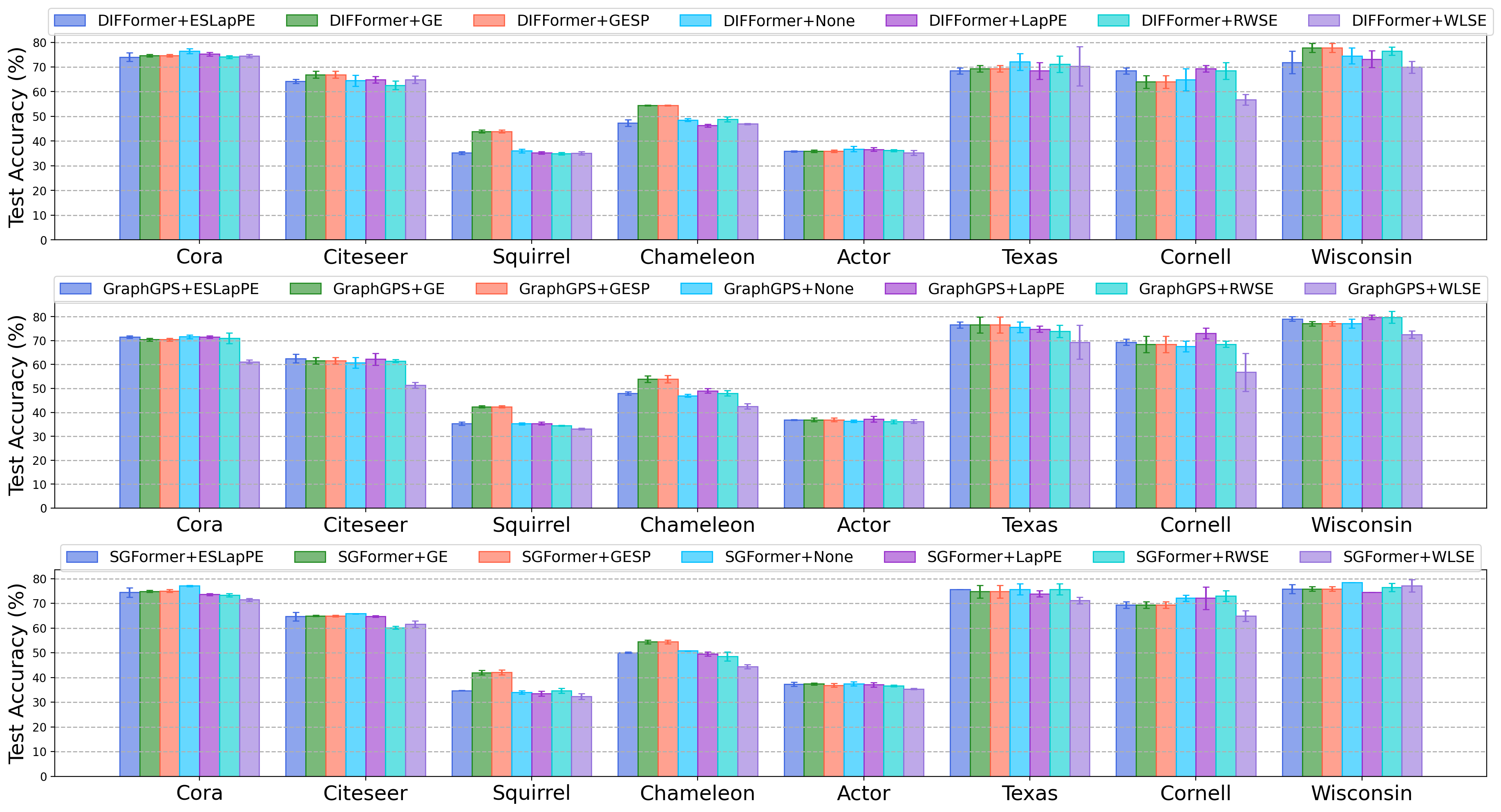}
    \caption{Accuracy results different models integrated with different positional encodings.}
    \label{pe_fig}
\end{figure}

\subsection{Positional Encodings Comparison}
\textbf{Observation 5: Effectiveness of degree-based positional encodings in dense graphs.} 

In dense graph settings such as \textit{Chameleon} and \textit{Squirrel}, degree-based positional encodings contribute significantly to model performance (see Figure~\ref{pe_fig}). We observe that both the simplified Graphormer Positional Encoder (GE), which uses only node degree embeddings, and the full version of Graphormer Positional Encoder (GESP), which additionally incorporates shortest path information, enhance model effectiveness. Interestingly, GE achieves performance comparable to GESP with reduced computational overhead.

This phenomenon is consistent with our intuition that in dense graphs, the average node degree is higher, leading to more distinctive degree distributions across nodes. This enables the model to encode coarse structural information without relying on more expensive global computations. Additionally, degree information is readily available and easy to integrate, making it a practical and effective component of positional encoding in such scenarios.
Furthermore, the utility of other positional encodings varies depending on the specific dataset and model architecture, suggesting that the choice of encoding should be made with task context in mind.

\textbf{Observation 6: Preprocessing Overhead of Positional Encodings on Large Graphs.} 
We observe that computing the preprocessing information required for various positional encodings can be significantly more time-consuming than model training, particularly on large graphs such as \textit{PubMed}. For instance, extracting random walk statistics takes approximately 10 minutes, spectral decomposition requires about one hour, and computing all-pairs shortest paths (including edge sequences) can take up to two hours. 

This discrepancy is largely attributable to the high time and space complexity of these operations. Shortest path computations between all pairs of vertices often require $O(n^3)$ time for dense representations, where $n$ is the number of nodes, while recording the edge attributes on the path requires $O(n^3)$ space. Spectral methods involve eigendecomposition of the graph Laplacian, typically requiring $O(n^3)$ time in the worst case, although efficient approximations can reduce this in practice. Similarly, random walk-based features depend on repeated matrix multiplications or sampling processes, which can be costly on large graphs.

These findings highlight a critical trade-off: while sophisticated positional encodings may enhance performance, their substantial preprocessing overhead may limit their applicability in time-sensitive or resource-constrained scenarios. Efficient approximations or adaptive use of encodings based on graph size and structure may be necessary to balance performance with scalability.




\section{Future Directions} 

Grounded in our empirical study, we outline several key research directions that merit particular attention for the future development of Graph Transformers. These directions reflect consistent trends observed across diverse datasets and architectures.

\textbf{Develop task-adapted architectures that distinguish between node-level and graph-level learning.} Our results show that models designed primarily for node-level tasks do not always transfer effectively to graph-level tasks. Future efforts may focus on clearer architectural separation of global graph representation modules and localized message-passing components, as well as careful choice of readout functions and normalization layers suited for the specific prediction target.

\textbf{Reduce the computational overhead of positional encoding preprocessing.} While positional encodings can enhance model performance, we find that generating required structural features (e.g., shortest paths, spectral components) can be time-intensive for large graphs. Future work may consider approximating these features through lighter heuristics, sampling-based estimation, or incorporating trainable positional encodings that can be optimized jointly with the model. These approaches may improve practicality without compromising performance.

\textbf{Tailor positional encoding strategies to graph properties and task requirements.} Our experiments indicate that the effectiveness of a given positional encoding—such as degree, spectral, or shortest path-based encodings—varies by dataset and model architecture. For example, degree-based encodings are more beneficial in dense graphs. Rather than applying a fixed encoding scheme across all tasks, future models could include configurable encoding pipelines or lightweight selection mechanisms informed by dataset statistics.
    
\textbf{Explore efficient attention mechanisms.} Our findings suggest that graph partition-based approaches offer a promising strategy for reducing computational overhead while preserving model expressiveness. Future work may further develop and refine such methods to enhance scalability on large graphs without compromising performance.

\textbf{Towards graph foundation model with GTs.} Through our comprehensive benchmark, we have revealed both the strengths and limitations of GTs. Notably, the ability of GTs to handle both homophily and heterophily, as well as their performance across various types of tasks, leads us to believe in the tremendous potential of GTs as graph foundation models. While transformer-based foundation models have already achieved great success in other domains such as computer vision and natural language processing, overcoming the current limitations of GTs and building foundation models based on GTs remains an exciting and promising research direction.

\section{Conclusions and Future Work} 

In this work, we present a comprehensive empirical study of Graph Transformers (GTs) by developing a unified evaluation framework. Our platform standardizes implementation details and experimental settings, allowing for fair and reproducible comparisons across a diverse set of state-of-the-art GTs and traditional Graph Neural Networks (GNNs). We benchmarked 16 models across a wide range of datasets that differ in task level, graph homophily, size, and sparsity. Our evaluation reveals several key findings about the impact of model architecture, attention design, positional encoding, and computational efficiency. These insights highlight both the strengths and limitations of current GTs and suggest practical considerations for model deployment. We also propose several potential directions based on our observations. We hope this work will serve as a solid foundation for rigorous, extensible, and transparent evaluation in future graph learning research.

\textbf{Broader Impacts and Limitations.} 

Although OpenGT has established a comprehensive benchmark, it also has some limitations that we aim to address in future work. 
First, we plan to expand the benchmark by incorporating a more diverse range of datasets to better evaluate methods across various scenarios.
Second, we aim to integrate additional GT implementations to provide a clearer picture of methodological advancements in the field. To maintain relevance with ongoing research, we commit to regular repository updates that align with emerging developments. 
Furthermore, we actively welcome constructive feedback and collaborative contributions to enhance both the practical utility and scientific rigor of our benchmarking framework.

\clearpage

\bibliographystyle{plain}
{
	\small
\bibliography{ref}
}

\newpage
\appendix

\section{Additional Results}

\subsection{Experiment results for different models}

In section 4, we investigated the performance of GTs and GNNs on different datasets. Table~\ref{tab:model-acc} illustrates the full experiment results for these models on node level datasets, and table~\ref{tab:model-graph} shows the full experiment results for these models on graph level datasets. In node level datasets, the numbers represent the accuracy of the corresponding experiment. In graph level datasets have different metrics according to their tasks. "OOM" stands for out of memory error in the corresponding experiment.

\begin{table}[h!]
  \caption{Model Accuracy Results on Node Level Datasets}
  \label{tab:model-acc}
\centering
\resizebox{\textwidth}{!}{%
\begin{tabular}{cccccccccc}
\toprule
Model/Dataset    & Cora          & Citeseer      & Pubmed        & Chameleon     & Squirrel      & Actor         & Cornell       & Texas         & Wisconsin     \\
\midrule
GCN              & 0.7180\scriptsize{±}0.0298  & 0.6173\scriptsize{±}0.0177 & 0.7150\scriptsize{±}0.0297  & 0.4415\scriptsize{±}0.0169 & 0.2917\scriptsize{±}0.0094 & 0.2713\scriptsize{±}0.0089 & 0.3243\scriptsize{±}0.0441 & 0.4505\scriptsize{±}0.0709 & 0.4379\scriptsize{±}0.0515 \\
GAT              & 0.7460\scriptsize{±}0.0116  & 0.6123\scriptsize{±}0.0162 & 0.7590\scriptsize{±}0.0184  & 0.4613\scriptsize{±}0.0021 & 0.3016\scriptsize{±}0.0086 & 0.2842\scriptsize{±}0.0121 & 0.3513\scriptsize{±}0.0441 & 0.3604\scriptsize{±}0.1469 & 0.3791\scriptsize{±}0.0606 \\
APPNP            & 0.7713\scriptsize{±}0.0101 & 0.6617\scriptsize{±}0.0266 & 0.7750\scriptsize{±}0.0046  & 0.4956\scriptsize{±}0.0100   & 0.3333\scriptsize{±}0.0088 & 0.2798\scriptsize{±}0.0017 & 0.4414\scriptsize{±}0.0127 & 0.5405\scriptsize{±}0.0000    & 0.4641\scriptsize{±}0.0092 \\
Graphtransformer & 0.7913\scriptsize{±}0.0052 & 0.6717\scriptsize{±}0.0062 & 0.7420\scriptsize{±}0.0099  & 0.4006\scriptsize{±}0.0055 & 0.2767\scriptsize{±}0.0077 & 0.2873\scriptsize{±}0.0006 & 0.3874\scriptsize{±}0.0127 & 0.6036\scriptsize{±}0.0255 & 0.5229\scriptsize{±}0.0462 \\
Graphormer       & 0.4323\scriptsize{±}0.0455 & 0.4360\scriptsize{±}0.0140   & OOM           & 0.5453\scriptsize{±}0.0136 & 0.4310\scriptsize{±}0.0120   & 0.3713\scriptsize{±}0.0057 & 0.7117\scriptsize{±}0.0127 & 0.7748\scriptsize{±}0.0127 & 0.7582\scriptsize{±}0.0244 \\
SAN              & 0.5403\scriptsize{±}0.0540  & OOM           & OOM           & 0.4189\scriptsize{±}0.0267 & OOM           & OOM           & 0.6667\scriptsize{±}0.0337 & 0.6667\scriptsize{±}0.0337 & 0.7843\scriptsize{±}0.0160  \\
GPS              & 0.7150\scriptsize{±}0.0051  & 0.6227\scriptsize{±}0.0246 & 0.7327\scriptsize{±}0.0184 & 0.4905\scriptsize{±}0.0092 & 0.3535\scriptsize{±}0.0062 & 0.3721\scriptsize{±}0.0117 & 0.7297\scriptsize{±}0.0221 & 0.7478\scriptsize{±}0.0127 & 0.7974\scriptsize{±}0.0092 \\
DIFFormer        & 0.7650\scriptsize{±}0.0099  & 0.6447\scriptsize{±}0.0217 & 0.7587\scriptsize{±}0.0066 & 0.4854\scriptsize{±}0.0058 & 0.3602\scriptsize{±}0.0070  & 0.3682\scriptsize{±}0.0110  & 0.6486\scriptsize{±}0.0441 & 0.7207\scriptsize{±}0.0337 & 0.7451\scriptsize{±}0.0320  \\
SpecFormer       & 0.4883\scriptsize{±}0.0084 & 0.4540\scriptsize{±}0.0100    & 0.6973\scriptsize{±}0.0079 & 0.6623\scriptsize{±}0.0089 & 0.4454\scriptsize{±}0.0141 & 0.3647\scriptsize{±}0.0087 & 0.7478\scriptsize{±}0.0255 & 0.7838\scriptsize{±}0.0221 & 0.7778\scriptsize{±}0.0092 \\
Exphormer        & 0.6120\scriptsize{±}0.0213  & 0.5117\scriptsize{±}0.0070  & 0.6987\scriptsize{±}0.0170  & 0.5161\scriptsize{±}0.0068 & 0.3490\scriptsize{±}0.0071  & 0.3577\scriptsize{±}0.0045 & 0.5405\scriptsize{±}0.0441 & 0.3784\scriptsize{±}0.0221 & 0.6994\scriptsize{±}0.0333 \\
GRIT             & OOM           & OOM           & OOM           & 0.4722\scriptsize{±}0.0260  & OOM           & OOM           & 0.6486\scriptsize{±}0.0584 & 0.7027\scriptsize{±}0.0382 & 0.8039\scriptsize{±}0.0160  \\
NodeFormer       & 0.7103\scriptsize{±}0.0175 & 0.5903\scriptsize{±}0.0204 & 0.6953\scriptsize{±}0.0144 & 0.4788\scriptsize{±}0.0058 & 0.3442\scriptsize{±}0.0063 & 0.3520\scriptsize{±}0.0047  & 0.6577\scriptsize{±}0.0459 & 0.6937\scriptsize{±}0.0255 & 0.7386\scriptsize{±}0.0403 \\
CoBFormer        & 0.7243\scriptsize{±}0.0118 & 0.6300\scriptsize{±}0.0127   & 0.7203\scriptsize{±}0.0133 & 0.5124\scriptsize{±}0.0057 & 0.3737\scriptsize{±}0.0049 & 0.3719\scriptsize{±}0.0088 & 0.7117\scriptsize{±}0.0337 & 0.7387\scriptsize{±}0.0637 & 0.7647\scriptsize{±}0.0320  \\
SGFormer         & 0.7703\scriptsize{±}0.0021 & 0.6580\scriptsize{±}0.0008  & 0.7347\scriptsize{±}0.0034 & 0.5080\scriptsize{±}0.0010   & 0.3401\scriptsize{±}0.0068 & 0.3746\scriptsize{±}0.0084 & 0.7207\scriptsize{±}0.0127 & 0.7568\scriptsize{±}0.0221 & 0.7843\scriptsize{±}0.0000    \\
DeGTA            & 0.7533\scriptsize{±}0.0056 & 0.6190\scriptsize{±}0.0232  & OOM           & 0.4920\scriptsize{±}0.0045  & 0.3164\scriptsize{±}0.0079 & 0.3507\scriptsize{±}0.0067 & 0.5315\scriptsize{±}0.0459 & 0.6126\scriptsize{±}0.0127 & 0.5229\scriptsize{±}0.0092 \\
\bottomrule
\end{tabular}%
}
\end{table}

\begin{table}[h!]
\centering
\caption{Model Results on Graph Level Datasets}
\label{tab:model-graph}
\resizebox{\textwidth}{!}{%
\begin{tabular}{ccccccc}
\toprule
Model/Dataset(Metric) & OGBG-MolHIV(AUC↑) & OGBG-MolPCBA(AP↑) & Peptides-Func(AP↑) & Peptides-Struct(MAE↓) & ZINC(MAE↓)    &  \\
\midrule
GCN              & 0.6888\scriptsize{±}0.0127 & 0.0953\scriptsize{±}0.0012 & 0.3614\scriptsize{±}0.0036 & 0.4347\scriptsize{±}0.0033 & 0.6090\scriptsize{±}0.0155  &  \\
GAT              & 0.7385\scriptsize{±}0.0300   & 0.1012\scriptsize{±}0.0015 & 0.3719\scriptsize{±}0.0041 & 0.4231\scriptsize{±}0.0049 & 0.6225\scriptsize{±}0.0096 &  \\
APPNP            & 0.7550\scriptsize{±}0.0050   & 0.0973\scriptsize{±}0.0011 & 0.3964\scriptsize{±}0.0022 & 0.4374\scriptsize{±}0.0050  & 0.6619\scriptsize{±}0.0067 &  \\
Graphtransformer & 0.6350\scriptsize{±}0.0112  & 0.0968\scriptsize{±}0.0000    & 0.3622\scriptsize{±}0.0036           & 0.4345\scriptsize{±}0.0026           & 0.6038\scriptsize{±}0.0151 &  \\
Graphormer       & OOM           & OOM           & OOM           & OOM           & 0.1305\scriptsize{±}0.0072 &  \\
SAN              & OOM           & OOM           & OOM           & OOM           & 0.1341\scriptsize{±}0.0063 &  \\
GPS              & 0.7715\scriptsize{±}0.0135 & OOM           & 0.6565\scriptsize{±}0.0038 & 0.2512\scriptsize{±}0.0008 & 0.1968\scriptsize{±}0.0120  &  \\
GPS+RWSE         & 0.7713\scriptsize{±}0.0105 & 0.2888\scriptsize{±}0.0019 & 0.6427\scriptsize{±}0.0035 & 0.2851\scriptsize{±}0.0112 & 0.0681\scriptsize{±}0.0004 &  \\
GPS+GE           & OOM           & OOM           & OOM           & OOM           & 0.1789\scriptsize{±}0.0071 &  \\
DIFFormer        & 0.7169\scriptsize{±}0.0103 & 0.0785\scriptsize{±}0.0021 & 0.3810\scriptsize{±}0.0013  & 0.4541\scriptsize{±}0.0005 & 0.6602\scriptsize{±}0.0294 &  \\
Exphormer        & OOM           & OOM           & 0.4003\scriptsize{±}0.0058 & 0.3297\scriptsize{±}0.0085 & 0.1905\scriptsize{±}0.0276 &  \\
GRIT             & 0.7608\scriptsize{±}0.0165 & 0.1931\scriptsize{±}0.0018 & OOM           & OOM           & 0.0607\scriptsize{±}0.0000    &  \\
GraphMLPMixer         & 0.6889\scriptsize{±}0.0146     & 0.1386\scriptsize{±}0.0018     & 0.4225\scriptsize{±}0.0095      & 0.3065\scriptsize{±}0.0019         & 0.3617\scriptsize{±}0.0222 &  \\
SGFormer         & 0.6794\scriptsize{±}0.0076 & 0.0933\scriptsize{±}0.0040  & 0.3726\scriptsize{±}0.0018 & 0.4492\scriptsize{±}0.0002 & 0.6703\scriptsize{±}0.0033 & \\
\bottomrule
\end{tabular}%
}
\end{table}

\newpage
\subsection{Experiment results for positional encodings}

In section 4, we also investigated the effect of different positional encodings inserted into different models.
Table~\ref{tab:pe-acc}, shows the accuracy of the models on selected datasets.

\begin{table}[h!]
\centering
\caption{Models with Positional Encoding Accuracy on Node Level Datasets}
\label{tab:pe-acc}
\resizebox{\textwidth}{!}{%
\begin{tabular}{ccccccccc}
\toprule
Model/Dataset    & Cora          & Citeseer      & Chameleon     & Squirrel      & Actor         & Cornell       & Texas         & Wisconsin     \\
\midrule
DIFFormer+None   & 0.7650\scriptsize{±}0.0099  & 0.6447\scriptsize{±}0.0217 & 0.4854\scriptsize{±}0.0058 & 0.3602\scriptsize{±}0.0070  & 0.3682\scriptsize{±}0.0110  & 0.6486\scriptsize{±}0.0441 & 0.7207\scriptsize{±}0.0337 & 0.7451\scriptsize{±}0.0320  \\
DIFFormer+LapPE  & 0.7523\scriptsize{±}0.0074 & 0.6490\scriptsize{±}0.0131  & 0.4627\scriptsize{±}0.0062 & 0.3529\scriptsize{±}0.0047 & 0.3667\scriptsize{±}0.0077 & 0.6937\scriptsize{±}0.0127 & 0.6847\scriptsize{±}0.0337 & 0.7320\scriptsize{±}0.0333  \\
DIFFormer+ESLapPE & 0.7403\scriptsize{±}0.0178 & 0.6420\scriptsize{±}0.0090 & 0.4737\scriptsize{±}0.0129 & 0.3519\scriptsize{±}0.0058 & 0.3592\scriptsize{±}0.0025 & 0.6847\scriptsize{±}0.0127 & 0.6847\scriptsize{±}0.0127 & 0.7189\scriptsize{±}0.0462 \\
DIFFormer+RWSE   & 0.7407\scriptsize{±}0.0063 & 0.6257\scriptsize{±}0.0173 & 0.4883\scriptsize{±}0.0105 & 0.3493\scriptsize{±}0.0043 & 0.3621\scriptsize{±}0.0045 & 0.6847\scriptsize{±}0.0337 & 0.7117\scriptsize{±}0.0337 & 0.7647\scriptsize{±}0.0160  \\
DIFFormer+GE     & 0.7467\scriptsize{±}0.0046 & 0.6693\scriptsize{±}0.0136 & 0.5446\scriptsize{±}0.0010  & 0.4393\scriptsize{±}0.0052 & 0.3590\scriptsize{±}0.0050   & 0.6396\scriptsize{±}0.0255 & 0.6937\scriptsize{±}0.0127 & 0.7778\scriptsize{±}0.0185 \\
DIFFormer+GESP   & 0.7467\scriptsize{±}0.0046 & 0.6693\scriptsize{±}0.0136 & 0.5446\scriptsize{±}0.0010  & 0.4393\scriptsize{±}0.0052 & 0.3590\scriptsize{±}0.0050   & 0.6396\scriptsize{±}0.0255 & 0.6937\scriptsize{±}0.0127 & 0.7778\scriptsize{±}0.0185 \\
DIFFormer+WLSE   & 0.7447\scriptsize{±}0.0061 & 0.6490\scriptsize{±}0.0149  & 0.4693\scriptsize{±}0.0031 & 0.3516\scriptsize{±}0.0067 & 0.3526\scriptsize{±}0.0106 & 0.5676\scriptsize{±}0.0221 & 0.7027\scriptsize{±}0.0796 & 0.6994\scriptsize{±}0.0244 \\
GPS+None         & 0.7160\scriptsize{±}0.0080   & 0.6077\scriptsize{±}0.0221 & 0.4700\scriptsize{±}0.0063   & 0.3525\scriptsize{±}0.0049 & 0.3632\scriptsize{±}0.0054 & 0.6757\scriptsize{±}0.0221 & 0.7568\scriptsize{±}0.0221 & 0.7712\scriptsize{±}0.0185 \\
GPS+LapPE        & 0.7150\scriptsize{±}0.0051  & 0.6227\scriptsize{±}0.0246 & 0.4905\scriptsize{±}0.0092 & 0.3535\scriptsize{±}0.0062 & 0.3721\scriptsize{±}0.0117 & 0.7297\scriptsize{±}0.0221 & 0.7478\scriptsize{±}0.0127 & 0.7974\scriptsize{±}0.0092 \\
GPS+ESLapPE      & 0.7150\scriptsize{±}0.0053  & 0.6253\scriptsize{±}0.0177 & 0.4788\scriptsize{±}0.0068 & 0.3529\scriptsize{±}0.0067 & 0.3689\scriptsize{±}0.0020  & 0.6937\scriptsize{±}0.0127 & 0.7658\scriptsize{±}0.0127 & 0.7909\scriptsize{±}0.0092 \\
GPS+RWSE         & 0.7097\scriptsize{±}0.0227 & 0.6147\scriptsize{±}0.0057 & 0.4803\scriptsize{±}0.0107 & 0.3445\scriptsize{±}0.0012 & 0.3614\scriptsize{±}0.0080  & 0.6847\scriptsize{±}0.0127 & 0.7387\scriptsize{±}0.0255 & 0.7974\scriptsize{±}0.0244 \\
GPS+GE           & 0.7040\scriptsize{±}0.0057  & 0.6157\scriptsize{±}0.0134 & 0.5395\scriptsize{±}0.0142 & 0.4236\scriptsize{±}0.0041 & 0.3691\scriptsize{±}0.0078 & 0.6847\scriptsize{±}0.0337 & 0.7658\scriptsize{±}0.0337 & 0.7712\scriptsize{±}0.0092 \\
GPS+GESP         & 0.7040\scriptsize{±}0.0057  & 0.6157\scriptsize{±}0.0134 & 0.5387\scriptsize{±}0.0152 & 0.4236\scriptsize{±}0.0041 & 0.3691\scriptsize{±}0.0078 & 0.6847\scriptsize{±}0.0337 & 0.7658\scriptsize{±}0.0337 & 0.7712\scriptsize{±}0.0092 \\
GPS+WLSE         & 0.6117\scriptsize{±}0.0079 & 0.5137\scriptsize{±}0.0118 & 0.4254\scriptsize{±}0.0107 & 0.3305\scriptsize{±}0.0036 & 0.3625\scriptsize{±}0.0078 & 0.5676\scriptsize{±}0.0796 & 0.6937\scriptsize{±}0.0709 & 0.7255\scriptsize{±}0.0160  \\
SGFormer+None    & 0.7703\scriptsize{±}0.0021 & 0.6580\scriptsize{±}0.0008  & 0.5080\scriptsize{±}0.0010   & 0.3401\scriptsize{±}0.0068 & 0.3746\scriptsize{±}0.0084 & 0.7207\scriptsize{±}0.0127 & 0.7568\scriptsize{±}0.0221 & 0.7843\scriptsize{±}0.0000    \\
SGFormer+LapPE   & 0.7357\scriptsize{±}0.0046 & 0.6473\scriptsize{±}0.0037 & 0.4949\scriptsize{±}0.0088 & 0.3346\scriptsize{±}0.0101 & 0.3706\scriptsize{±}0.0089 & 0.7207\scriptsize{±}0.0459 & 0.7387\scriptsize{±}0.0127 & 0.7451\scriptsize{±}0.0000    \\
SGFormer+ESLapPE & 0.7440\scriptsize{±}0.0191  & 0.6467\scriptsize{±}0.0175 & 0.5007\scriptsize{±}0.0027 & 0.3468\scriptsize{±}0.0008 & 0.3728\scriptsize{±}0.0089 & 0.6937\scriptsize{±}0.0127 & 0.7568\scriptsize{±}0.0000    & 0.7582\scriptsize{±}0.0185 \\
SGFormer+RWSE    & 0.7323\scriptsize{±}0.0066 & 0.6017\scriptsize{±}0.0057 & 0.4854\scriptsize{±}0.0183 & 0.3465\scriptsize{±}0.0098 & 0.3662\scriptsize{±}0.0036 & 0.7297\scriptsize{±}0.0221 & 0.7568\scriptsize{±}0.0221 & 0.7647\scriptsize{±}0.0160  \\
SGFormer+GE      & 0.7487\scriptsize{±}0.0047 & 0.6493\scriptsize{±}0.0026 & 0.5446\scriptsize{±}0.0075 & 0.4198\scriptsize{±}0.0095 & 0.3737\scriptsize{±}0.0038 & 0.6937\scriptsize{±}0.0127 & 0.7478\scriptsize{±}0.0255 & 0.7582\scriptsize{±}0.0092 \\
SGFormer+GESP    & 0.7500\scriptsize{±}0.0059   & 0.6490\scriptsize{±}0.0028  & 0.5446\scriptsize{±}0.0075 & 0.4201\scriptsize{±}0.0101 & 0.3684\scriptsize{±}0.0075 & 0.6937\scriptsize{±}0.0127 & 0.7478\scriptsize{±}0.0255 & 0.7582\scriptsize{±}0.0092 \\
SGFormer+WLSE    & 0.7147\scriptsize{±}0.0058 & 0.6160\scriptsize{±}0.0127  & 0.4437\scriptsize{±}0.0085 & 0.3231\scriptsize{±}0.0114 & 0.3535\scriptsize{±}0.0030  & 0.6486\scriptsize{±}0.0221 & 0.7117\scriptsize{±}0.0127 & 0.7712\scriptsize{±}0.0244 \\
\bottomrule
\end{tabular}%
}
\end{table}

\subsection{Experiment results for time efficiency}
Table~\ref{tab:model-time} demonstrates the average time before reaching the best validation performance on node level datasets, and table~\ref{tab:model-time-graph} represents the average time before reaching the best validation performance on graph level datasets.

\begin{table}[h!]
\centering
\caption{Time Cost (in seconds) for Models on Node Level Datasets}
\label{tab:model-time}
\resizebox{\textwidth}{!}{%
\begin{tabular}{cccccccccc}
\toprule
Model/Dataset & Cora            & Citeseer       & Pubmed          & Chameleon      & Squirrel       & Actor           & Cornell       & Texas          & Wisconsin     \\
\midrule
GCN           & 4.6758\scriptsize{±}1.8777   & 1.5058\scriptsize{±}0.6589  & 2.1792\scriptsize{±}1.0022   & 1.2764\scriptsize{±}0.8685  & 2.7262\scriptsize{±}2.7092  & 2.1152\scriptsize{±}1.3510    & 1.1820\scriptsize{±}1.4025  & 2.5338\scriptsize{±}2.5156  & 0.6411\scriptsize{±}0.6405 \\
GAT           & 1.0454\scriptsize{±}0.6715   & 2.3343\scriptsize{±}1.4398  & 1.7172\scriptsize{±}1.0120    & 1.0557\scriptsize{±}0.6939  & 1.5134\scriptsize{±}0.8319  & 1.0996\scriptsize{±}0.7380    & 0.6441\scriptsize{±}0.6057 & 0.7580\scriptsize{±}0.8831   & 0.8363\scriptsize{±}0.8772 \\
APPNP         & 1.5151\scriptsize{±}1.1574   & 1.4387\scriptsize{±}0.8186  & 2.7982\scriptsize{±}0.7725   & 2.0179\scriptsize{±}0.7011  & 3.4525\scriptsize{±}0.8295  & 5.7091\scriptsize{±}3.2671   & 2.0465\scriptsize{±}0.4733 & 1.2664\scriptsize{±}0.6156  & 1.1840\scriptsize{±}0.7968  \\
Graphtransformer & 1.4213\scriptsize{±}0.7181 & 1.6576\scriptsize{±}0.6223 & 4.5835\scriptsize{±}2.7363 & 0.8653\scriptsize{±}0.6104 & 1.6104\scriptsize{±}0.6065 & 1.4447\scriptsize{±}0.7919 & 0.5144\scriptsize{±}0.6474 & 0.9030\scriptsize{±}0.8468 & 0.5822\scriptsize{±}0.5227 \\
Graphormer    & 6.8356\scriptsize{±}1.9709   & 16.4166\scriptsize{±}7.6135 & OOM             & 6.4411\scriptsize{±}3.6455  & 20.2078\scriptsize{±}4.0263 & 47.1688\scriptsize{±}11.1344 & 1.3133\scriptsize{±}0.5959 & 1.2481\scriptsize{±}0.5990   & 1.6568\scriptsize{±}0.5251 \\
SAN           & 55.3741\scriptsize{±}18.4253 & OOM            & OOM             & 26.2862\scriptsize{±}8.7625 & OOM            & OOM             & 2.1613\scriptsize{±}1.5735 & 12.6906\scriptsize{±}3.4535 & 1.6650\scriptsize{±}0.6405  \\
GPS           & 3.9762\scriptsize{±}2.0044   & 5.3102\scriptsize{±}1.8808  & 4.7728\scriptsize{±}1.5635   & 1.4809\scriptsize{±}0.5850   & 2.3029\scriptsize{±}0.5490   & 5.0642\scriptsize{±}1.7482   & 1.0920\scriptsize{±}0.8521  & 0.8722\scriptsize{±}0.6578  & 0.9051\scriptsize{±}0.6588 \\
DIFFormer     & 8.8231\scriptsize{±}2.4739   & 7.4126\scriptsize{±}1.6411  & 8.6028\scriptsize{±}1.4446   & 7.5940\scriptsize{±}5.2168   & 7.9667\scriptsize{±}0.7576  & 11.3840\scriptsize{±}1.7681   & 2.4829\scriptsize{±}0.5649 & 2.5257\scriptsize{±}0.7128  & 1.9260\scriptsize{±}0.6574  \\
SpecFormer    & 2.2153\scriptsize{±}1.6275   & 2.0729\scriptsize{±}0.7600    & 77.6776\scriptsize{±}44.3504 & 4.3569\scriptsize{±}0.7850   & 30.7572\scriptsize{±}2.4627 & 5.7146\scriptsize{±}1.4039   & 1.0163\scriptsize{±}0.5152 & 1.4125\scriptsize{±}0.7841  & 0.8985\scriptsize{±}0.6040  \\
Exphormer     & 2.8879\scriptsize{±}3.7204   & 0.8047\scriptsize{±}0.6312  & 1.4392\scriptsize{±}0.5586   & 0.7637\scriptsize{±}0.6025  & 0.8945\scriptsize{±}0.6188  & 0.8848\scriptsize{±}0.6971   & 0.6404\scriptsize{±}0.5617 & 0.5371\scriptsize{±}0.5906  & 0.5894\scriptsize{±}0.6335 \\
GRIT          & OOM             & OOM            & OOM             & 13.6541\scriptsize{±}3.6749 & OOM            & OOM             & 1.3395\scriptsize{±}0.4961 & 1.4999\scriptsize{±}0.4825  & 1.3850\scriptsize{±}0.8212  \\
NodeFormer    & 17.7103\scriptsize{±}1.0020   & 15.1504\scriptsize{±}6.2730  & 6.0357\scriptsize{±}1.5334   & 5.7416\scriptsize{±}1.0657  & 4.4503\scriptsize{±}0.2750   & 2.9424\scriptsize{±}0.4452   & 2.8189\scriptsize{±}1.1180  & 1.4983\scriptsize{±}0.7429  & 2.5140\scriptsize{±}1.7377  \\
CoBFormer     & 9.2926\scriptsize{±}1.4867   & 10.5013\scriptsize{±}2.5827 & 7.2540\scriptsize{±}6.3860     & 1.5291\scriptsize{±}0.8226  & 2.9565\scriptsize{±}0.8684  & 1.8607\scriptsize{±}0.9457   & 1.3847\scriptsize{±}0.7391 & 1.6133\scriptsize{±}0.5146  & 1.0630\scriptsize{±}0.6563  \\
SGFormer      & 1.9826\scriptsize{±}0.6895   & 1.9616\scriptsize{±}0.7186  & 6.7971\scriptsize{±}1.6985   & 4.2802\scriptsize{±}0.4252  & 3.1163\scriptsize{±}0.5310   & 3.1916\scriptsize{±}0.7963   & 4.3429\scriptsize{±}1.0589 & 2.2112\scriptsize{±}0.5968  & 3.7581\scriptsize{±}0.7719 \\
DeGTA         & 3.4608\scriptsize{±}0.0845   & 4.1701\scriptsize{±}1.2229  & OOM             & 1.9460\scriptsize{±}0.8257   & 3.3788\scriptsize{±}0.7975  & 4.9141\scriptsize{±}1.0583   & 1.9555\scriptsize{±}1.1011 & 1.2336\scriptsize{±}0.7317  & 2.1124\scriptsize{±}0.5427 \\
\bottomrule
\end{tabular}%
}
\end{table}

\begin{table}[h!]
\centering
\caption{Time Cost (in seconds) for Models on Graph Level Datasets}
\label{tab:model-time-graph}
\resizebox{\textwidth}{!}{%
\begin{tabular}{cccccc}
\toprule
Model/Dataset    & OGBG-MolHIV         & OGBG-MolPCBA        & Peptides-Func      & Peptides-Struct     & ZINC                 \\
\midrule
GCN              & 617.6852\scriptsize{±}73.0724    & 6066.3818\scriptsize{±}511.9715  & 1270.1438\scriptsize{±}277.9431 & 463.8961\scriptsize{±}86.5786    & 3705.1342\scriptsize{±}2612.8606  \\
GAT              & 3582.2222\scriptsize{±}806.0562  & 6920.119\scriptsize{±}880.9262   & 1639.9405\scriptsize{±}167.5811 & 890.4499\scriptsize{±}126.3149   & 3677.9652\scriptsize{±}867.3893   \\
APPNP            & 3464.0672\scriptsize{±}954.0117  & 8557.5001\scriptsize{±}610.4127  & 1766.8020\scriptsize{±}426.0009  & 485.1932\scriptsize{±}33.5860     & 13088.5143\scriptsize{±}9262.9440  \\
Graphtransformer & 492.7499\scriptsize{±}122.2428   & 3788.1170\scriptsize{±}0.0000        & 427.5714\scriptsize{±}69.2988                & 219.3224\scriptsize{±}48.4751                 & 707.9019\scriptsize{±}79.8120      \\
Graphormer       & OOM                 & OOM                 & OOM                & OOM                 & 13848.4759\scriptsize{±}2600.9056 \\
SAN              & OOM                 & OOM                 & OOM                & OOM                 & 44432.3986\scriptsize{±}7142.7024 \\
GPS              & 3418.8905\scriptsize{±}1026.4501 & OOM                 & 1204.2498\scriptsize{±}147.9630  & 1007.4493\scriptsize{±}213.8006  & 28504.2386\scriptsize{±}4033.5811 \\
GPS+RWSE      & 2904.4005\scriptsize{±}594.5993  & 8570.1206\scriptsize{±}1917.9189 & 1688.6532\scriptsize{±}258.8618  & 312.5400\scriptsize{±}99.6745     & 34975.4642\scriptsize{±}4523.6709  \\
GPS+GE           & OOM                 & OOM                 & OOM                & OOM                 & 37495.6849\scriptsize{±}491.7593  \\
DIFFormer     & 2334.0867\scriptsize{±}1370.0275 & 9251.2273\scriptsize{±}3768.8335 & 3038.4863\scriptsize{±}1188.6412 & 2377.5684\scriptsize{±}782.3884 & 27983.3716\scriptsize{±}10179.6969 \\
Exphormer        & OOM                 & OOM                 & 189.0655\scriptsize{±}122.4873  & 3708.7887\scriptsize{±}3140.1501 & 19090.4587\scriptsize{±}3693.4416 \\
GRIT             & 4439.1861\scriptsize{±}3529.5704 & 15486.5944\scriptsize{±}321.1582 & OOM                & OOM                 & 55160.9429\scriptsize{±}0.0000       \\
GraphMLPMixer & 388.6281\scriptsize{±}48.4666    & 8629.184\scriptsize{±}1776.4543  & 249.1331\scriptsize{±}98.8619    & 216.3899\scriptsize{±}57.6617   & 4887.2565\scriptsize{±}1979.3262   \\
SGFormer      & 1506.3488\scriptsize{±}510.3602  & 9196.8431\scriptsize{±}1009.7321 & 2220.7436\scriptsize{±}308.8496  & 2097.2997\scriptsize{±}337.3081 & 2564.1735\scriptsize{±}710.1304  \\
\bottomrule
\end{tabular}%
}
\end{table}

\section {Hyperparameter Settings}

We perform grid search for the hyperparameters not specifically mentioned in the corresponding papers. The search space is shown in table ~\ref{tab:hyperparam}.

\begin{table}[h!]
\centering
\caption{Hyper-parameter search space of all implemented GT methods.}
\renewcommand\arraystretch{1.2}
\label{tab:hyperparam}
    \begin{tabular}{clc}
    \hline
    \textbf{Algorithm} & \textbf{Hyper-parameter} & \textbf{Search Space} \\ 
    \hline
    \hline
    \multirow{6}{*}{General Settings} & learning rate & 1e-4,3e-4,1e-3,3e-3,1e-2 \\  
     & weight decay &  1e-5,3e-5,1e-4,3e-4,1e-3 \\  
     & number of layers & 1, 2, 3, 4\\  
     & number of attention heads & 1,2,3,4 \\   
     & dropout & 0, 0.2, 0.5, 0.8 \\   
     & attention dropout & 0, 0.2, 0.5, 0.8 \\ 
     \hline
     
    \multirow{2}{*}{SGFormer~\cite{sgformer}}
     & aggregate method & add, cat\\  
     & graph weight & 0.2, 0.5, 0.8 \\ 
     \hline

    \multirow{1}{*}{DIFFormer~\cite{difformer}}
     & graph weight & 0.2, 0.5, 0.8 \\ 
     \hline
     
     \multirow{1}{*}{DeGTA~\cite{degta}}
     & $K$ & 2, 4, 8 \\ 
     \hline
    
    \end{tabular}
\end{table}

\section{Reproducibility}
\label{reproducibility}
All of OpenGT's experimental results are highly reproducible. We provide more detailed information on the following aspects to ensure the reproducibility of the experiments.

{\bf Accessibility.} You can access all datasets, algorithm implementations, and experimental configurations in our open source project \url{https://github.com/eaglelab-zju/OpenGT} without a personal request.

{\bf Dataset.} All datasets used are publicly available. 
The \textit{Cora}, \textit{Citeseer}, and \textit{Pubmed} dataset\citep{sen2008collective}s are accessible online and are used under the Creative Commons 4.0 license. 
The webpage networks \textit{Squirrel} and \textit{Chameleon} \citep{rozemberczki2019gemsec},
the \textit{Actor} \citep{shchur2018pitfalls} dataset, the university web page datasets \textit{Texas}, \textit{Cornell}, \textit{Wisconsin} \citep{pei2020geom}
and molecular graphs \textit{ZINC}\citep{irwin2012zinc} are public available in Pytorch-Geometric(PyG), which use MIT Liscence.
For two Open Graph Benchmark (OGB) datasets \textit{OGBG-MolHIV} and \textit{OGBG-MolPCBA} \citep{hu2020open}, they use MIT Licence.
For two peptide datasets, \textit{Peptides-func} and \textit{Peptides-struct} \citep{dwivedi2022long}, they use MIT Licence.
All of these datasets are for academic research and do not contain any personally identifiable information or offensive content.

{\bf Documentation and uses.} We've dedicated ourselves to providing users with comprehensive documentation, guaranteeing a smooth experience with our library. Our code includes ample comments to enhance readability. Furthermore, we furnish all essential files to replicate experimental outcomes, which also serve as illustrative guides on library utilization. Running the code is straightforward; users need only execute the '.sh' and '.py' files with provided config files and specified arguments like data, method, and GPU.

{\bf License.} We use an MIT license for our open-sourced project.

{\bf Code maintenance.} We are dedicated to maintaining our code through continuous updates, actively engaging with user feedback, and addressing any issues promptly. Additionally, we are eager to receive contributions from the community to improve our library and benchmark algorithms. However, we will uphold rigorous version control measures to uphold reproducibility standards during maintenance procedures.

\end{document}